\documentclass[11pt]{article}
\usepackage[margin=1in]{geometry}
\usepackage{amsmath, amsthm, amssymb, amsfonts,framed,subfigure,graphicx,float}
\usepackage{natbib,hyperref,color,stfloats,tabularx,enumitem}

\newtheorem{theorem}{Theorem}
\newtheorem{lemma}{Lemma}
\theoremstyle{definition}
\newtheorem{definition}{Definition}

\newcommand{\ddt}{\frac{\mathrm{d}}{\mathrm{d}t}}
\newcommand{\opt}{\mathrm{OPT}}
\newcommand{\mdp}{\mathcal{M}}
\newcommand{\states}{\mathcal{S}}
\newcommand{\actions}{\mathcal{A}}
\newcommand{\trans}{\mathcal{P}}

\newcommand{\expect}{\mathbb{E}}

\title{Planning with Submodular Objective Functions}
\author{
Ruosong Wang\thanks{Equal contribution} \\ Carnegie Mellon University\\ \texttt{ruosongw@andrew.cmu.edu}
\and
Hanrui Zhang$^*$ \\ Duke University\\ \texttt{hrzhang@cs.duke.edu}
\and
Devendra Singh Chaplot  \\ Carnegie Mellon University\\ \texttt{chaplot@cs.cmu.edu}
\and
Denis Garagi\'c \\ Sarcos Robotics \\ \texttt{d.garagic@sarcos.com}
\and
Ruslan Salakhutdinov \\ Carnegie Mellon University\\ \texttt{rsalakhu@cs.cmu.edu}
}
\date{}
\begin{document}
\maketitle

\begin{abstract}
We study planning with submodular objective functions, where instead of maximizing the cumulative reward, the goal is to maximize the objective value induced by a submodular function. Our framework subsumes standard planning and submodular maximization with cardinality constraints as special cases, and thus many practical applications can be naturally formulated within our framework. Based on the notion of multilinear extension, we propose a novel and theoretically principled algorithmic framework for planning with submodular objective functions, which recovers classical algorithms when applied to the two special cases mentioned above. Empirically, our approach significantly outperforms baseline algorithms on synthetic environments and navigation tasks.
\end{abstract}

\section{Introduction}
Modern reinforcement learning and planning algorithms have achieved tremendous successes on various tasks~\citep{mnih2015human, silver2017mastering}.
However, most of these algorithms work in the standard Markov decision process (MDP) framework where the goal is to maximize the cumulative reward and thus it can be difficult to apply them to 
 various practical sequential decision-making problems. 
In this paper, we study planning in generalized MDPs, where instead of maximizing the cumulative reward, the goal is to maximize the objective value induced by a submodular function. 

To motivate our approach, let us consider the following scenario: a company manufactures cars, and as part of its customer service, continuously monitors the status of all cars produced by the company.
Each car is equipped with a number of sensors, each of which constantly produces noisy measurements of some attribute of the car, e.g., speed, location, temperature, etc.
Due to bandwidth constraints, at any moment, each car may choose to transmit data generated by a single sensor to the company.
The goal is to combine the statistics collected over a fixed period of time, presumably from multiple sensors, to gather as much information about the car as possible.

Perhaps one seemingly natural strategy is to transmit only data generated by the most ``informative'' sensor.
However, as the output of a sensor remains the same between two samples, it is pointless to transmit the same data multiple times.
One may alternatively try to order sensors by their ``informativity'' and always choose the most informative sensor that has not yet transmitted data since the last sample was generated.
This strategy, however, is also suboptimal, since individual sensors provide diminishing marginal information gain --- the more we already know about the car, the less information an additional sensor provides.
In general, which sensors are more informative depends heavily on which other sensors have already transmitted data.
Choosing the set of sensors to transmit data is therefore a combinatorial optimization problem, where the amount of information provided is a submodular function of the set of sensors to transmit data.\footnote{Recall that Shannon entropy is monotone and submodular for discrete-valued random variables~\citep{fujishige1978polymatroidal}.}
In some cases, how informative a sensor is may even depend on the precise output of other sensors.
There, the problem can further be viewed as a generalized MDP\footnote{
An alternative approach is to formulate the problem as a Partially Observable Markov Decision Process (POMDP)~\citep{smallwood1973optimal}, which is a general framework that captures many adaptive optimization problems.
However, solving POMDPs is known to be PSPACE-Hard~\citep{papadimitriou1987complexity}, thus typically heuristic algorithms with no approximation guarantees are applied.~\citep{pineau2006anytime, ross2008online}
In this paper, we focus on theoretically principled algorithms and therefore do not employ such an approach. 
} --- in each step, we choose a sensor to transmit and observe the transmitted data, which takes us randomly to a new state, where we choose the next sensor to transmit data.
The main difference from traditional MDPs is that here, the objective function is a submodular function of the entire trajectory, rather than the cumulative reward from individual steps.
Indeed, research in the area of sensor selection~\citep{shamaiah2010greedy, kirchner2019heterogeneous} suggests that the problem of measurement selection does have the structure of submodular maximization, where the objective function is the log-determinant of the sum of Fisher information matrices associated with the measurements. 

As another example, in navigation tasks, the agent sequentially chooses actions to perform (changing speed or changing direction) to gain as much information as possible about the locations to be explored.
Here, how informative an action heavily depends on the navigation history of the agent, and it is not always possible to quantify the amount of information gained as a real reward value.
Consequently, for navigation tasks, a natural choice is to employ a submodular objective function which depends on the entire trajectory to measure the amount of information gained.

In this paper, we consider the above problems in a rather general sense.
Our goal is to answer the following question: how to act (approximately) optimally in planning problems, when the objective function is a submodular function of the trajectory?

\paragraph{Our Contributions.}
In this paper, we focus on planning in tabular MDPs with submodular objective functions, i.e., given a tabular MDP, our goal is to output a policy which approximately maximizes the submodular objective function.
Based on the notion of multilinear extension~\citep{calinescu2011maximizing} for submodular maximization, we design a theoretically principled  and efficient algorithm, which builds a random policy in a greedy fashion, by expanding gradually in the direction that maximally increases the value of a proxy of the reward function.
Theoretically, we give approximation guarantees for our algorithm when applied to general MDPs with submodular objective functions.
Moreover, we show that our algorithm, when combined with carefully designed rounding techniques, recovers classical algorithms when applied to two important special cases of the problem.
Empirically, we evaluate our algorithm on two environments: a synthetic environment and a navigation task based on the Matterport Dataset~\citep{chang2017matterport3d}.
Combined with our rounding techniques, our algorithm achieves better performance compared to various baseline algorithms, demonstrating the practicality of our approach.

\subsection{Related Work}\label{sec:rel}
\paragraph{Reinforcement Learning and Planning.} Most planning and reinforcement learning algorithms rely on the MDP framework.
For the tabular setting, there is a long line of work studying the optimal sample complexity and regret bound.
We refer interested readers to~\citep{kearns2002near,strehl2006pac,jaksch2010near,azar2017minimax,agrawal2017optimistic,sidford2018variance,sidford2018near,kakade2018variance, jin2018q} and references therein.
However, to our knowledge, all these works only study the case where the reward values are real numbers and the objective function is the cumulative reward, and thus cannot be applied to the setting studied in this paper where the objective function is a general submodular function.

\paragraph{Submodular Optimization.} Submodular optimization is an important topic in combinatorial optimization and has been extensively studied under various combinatorial constraints, including the knapsack constraint~\citep{sviridenko2004note, leskovec2007cost, badanidiyuru2014fast, ene2017nearly}, the matroid constraint~\citep{vondrak2008optimal, calinescu2011maximizing, filmus2014monotone, badanidiyuru2014fast, mirzasoleiman2015lazier, ene2019towards}, and the path constraint~\citep{chekuri2005recursive, singh2007efficient}.
We refer interested readers to the survey~\citep{krause2014submodular} for previous work on this topic.
However, as far as we are aware, there is no previous work studying submodular optimization under the general MDP framework that will be studied in this paper.

Our algorithmic framework is built upon the multilinear extension and continuous greedy framework introduced by~\citep{vondrak2008optimal, calinescu2011maximizing}.
In~\citep{vondrak2008optimal, calinescu2011maximizing}, continuous greedy was applied to solve submodular maximization under the matroid constraint. 
In this paper, we give the first efficient implementation and theoretical analysis of continuous greedy when applied to planning with 
submodular objective functions.

\paragraph{Sensor Selection and Its Generalizations. }
Sensor selection is an important topic in machine learning, robotics and signal processing (we refer interested readers to~\citep{joshi2008sensor} and references therein for prior works on this topic), and here we focus on prior works that generalize the sensor selection problem and are related to MDPs/POMDPs and submodular functions. 
\citet{golovin2011adaptive} studied the Adaptive Stochastic Optimization problem which can be formulated as a POMDP, and generalized the notion of submodularity to the adaptive setting. 
However, although the problem studied in~\citep{golovin2011adaptive} can be formulated as a special case of general POMDPs, their framework does not own the full generality of POMDPs. In particular, unlike our framework, the framework in~\citep{golovin2011adaptive} does not subsume standard MDPs as special cases.
Later,~\citet{vien2015touch} integrated an action hierarchy into the framework in~\citep{golovin2011adaptive}.
\citet{satsangi2015exploiting} modeled the dynamic sensor selection problem as a POMDP, and proposed an efficient algorithm based on greedy maximization and Point-Based Value Iteration (PBVI) by exploiting the submodularity of the value function. 
Compared to their approach, our framework does not assume the submodularity of the value function. Instead, we assume the objective function is a submodular function with respect to the visited state-action pairs. 
Moreover, unlike our framework, the framework in~\citep{satsangi2015exploiting} (which is a special case of the general POMDP framework) does not subsume standard MDPs as special cases.
\citet{greigarn2019task} considered a partially observable setting, and modeled the problem of sensing by separating the actions that only affect the observations (sensing actions) from those that affect only the transition between states (task actions).  They assume that a policy for task actions is given externally, and aim to find a good policy for sensing actions in order to minimize uncertainty about the current state.  In contrast, we consider a unified model where there is no explicit separation between ``sensing actions'' and ``task actions'', and we aim to find a good policy for the entire planning problem.
Moreover, \citet{greigarn2019task} focused on minimizing the conditional entropy, while in this work we consider general submodular objective functions.

\paragraph{Other Related Work. }
\citet{kumar2009event} considered planning under uncertainty for multiple agents and studied event-detecting multi-agent MDPs (eMMDPs). 
\citet{kumar2009event} proposed a constant-factor approximation algorithm for solving eMMDPs by exploiting the submodularity of the evaluation function of eMMDPs.
The objective function considered in~\citep{kumar2009event} is still the cumulative reward over steps, while in this paper we consider general submodular objective functions. 
\citet{kumar2017decentralized} considered decentralized stochastic planning for a team of agents and studied Transition Independent Dec-MDPs (TI-Dec-MDPs). 
\citet{kumar2017decentralized} focused on the case where the reward function is submodular and provided an efficient algorithm to solve TI-Dec-MDPs.
Although the reward function is assumed to be submodular, the objective function considered in~\citep{kumar2017decentralized} is still the cumulative reward.
The framework in~\citep{kumar2017decentralized} was later generalized to non-dedicated agents in~\citep{agrawal18decentralized}.
\section{Preliminaries}\label{sec:pre}
\paragraph{Notations.}
We write $[n]$ to denote the set $\{1,2,\ldots, n\}$.
For a given set $\Omega$, we use $2^{\Omega}$ to denote the power set of $S$, i.e., the set of all possible subsets of $\Omega$.
For two vectors $x, y \in \mathbb{R}^d$, we write $x \le y$ if $x_i \le y_i$ for all $i \in [d]$. We define $x \ge y$ analogously. 

We recall the definition of submodular functions and monotone functions.
\begin{definition}[Submodular Function]
    For a given set $\Omega$, a function $f: 2^{\Omega} \to \mathbb{R}_+$ is {\em submodular}, if for any $S, T \subseteq \Omega$, $f(S) + f(T) \ge f(S \cup T) + f(S \cap T)$.
\end{definition}
\begin{definition}[Monotone Function]
    For a given set $\Omega$, a function $f: 2^{\Omega} \to \mathbb{R}_+$ is {\em monotone}, if for any $S \subseteq T$, $f(S) \le f(T)$.
\end{definition}
\subsection{Episodic Planning with Submodular Objective Functions}
\label{sec:mdp}
In this section, we formally define the model that will be studied in this paper.

A {\em Markov decision process with submodular objective function} (submodular MDP) $\mdp$ is a tuple $\mdp =\left(\states, \actions, H,\trans, f \right)$.
Here, $\states$ is the state space, $\actions$ is the action space, $H \in \mathbb{Z}_+$ is the planning horizon, and $\trans: \states \times \actions \rightarrow \triangle(\states)$ is the transition operator which takes a state-action pair and returns a distribution over states.
For each $h \in [H]$, we use $\states_h \subseteq \states$ to denote the set of states at level $h$, and without loss of generality, we assume $\{\states_h\}_{h = 1}^H$ do not intersect with each other.\footnote{
In general, we may replace each state $s \in \states$ with a pair $(s, h) \in \states \times [H]$, which will only increase the size of the state space by a factor of $H$.
}
The objective function $f : 2^{\states \times \actions} \to \mathbb{R}_+$ maps a subset of state-action pairs to an objective value.
Here, we assume $f$ is monotone and submodular.

In this paper, we assume there is a fixed initial state $s_1$.\footnote{Note that this assumption is without loss of generality. If $s_1$ is sampled from an initial state distribution $\mathcal{D}$, then we may set $\trans(s_1,a) = \mathcal{D}$ for all $a \in \actions$ and now our $s_2$ is equivalent to the initial state sampled from $\mathcal{D}$.}
A policy $\pi: \states^* \rightarrow \triangle(\actions)$ prescribes a distribution over actions for a sequence of states.
Given a submodular MDP $\mdp$, a policy $\pi$ induces a (random) trajectory $s_1,a_1,s_2,a_2,s_3,a_3,\ldots,s_H,a_H$
where $a_1 \sim \pi(s_1)$, $s_2 \sim \trans(s_1,a_1)$, $a_2 \sim \pi(s_1, s_2)$, $s_3 \sim \trans(s_2, a_2)$, etc.
For a given trajectory, its objective value is given by $f(\{(s_1, a_1), (s_2, a_2), \ldots, (s_H, a_H)\})$ where $f : 2^{\states \times \actions} \to \mathbb{R}_+$ is the objective function mentioned above.

We focus on the planning problem in submodular MDP.
Given a submodular MDP $\mdp$, our goal is to find a policy $\pi$ that approximately maximizes the expected objective value $f(\pi) = \expect \left[f(\{(s_1, a_1), (s_2, a_2), \ldots, (s_H, a_H)\})\mid \pi\right]$.
We use $\pi^*$ to denote the optimal policy and $f(\pi^*)$ to denote its expected objective value.

\subsection{Special Cases}\label{sec:example}
In this section, we list a few special cases to demonstrate the generality of our model.

\paragraph{Standard MDP.} In the standard MDP setting, for each state-action pair $(s, a)$, there is a reward value $r(s, a) \in \mathbb{R}$ associated with that state-action pair, and the goal is to find a policy which maximizes the cumulative reward $\sum_{h = 1}^H r(s_h, a_h)$. The standard MDP setting can be readily formulated as a submodular MDP, simply by setting the objective function to be $f(\{(s_1, a_1), (s_2, a_2), \ldots, (s_H, a_H)\}) = \sum_{h = 1}^H r(s_h, a_h)$.

\paragraph{Cardinality Constraints.} Suppose we are given a finite set of elements $\Omega = \{\omega_1, \omega_2, \ldots, \omega_n\}$ with size $|\Omega| = n$, a monotone submodular function $\hat{f} : 2^{\Omega} \to \mathbb{R}_+$ and a cardinality constraint $k \le n$, in the {\em submodular maximization with cardinality constraint} problem, the goal is to find a subset $S \subseteq \Omega$ with  with cardinality $|S| = k$, such that $f(S)$ is maximized.
This is a classical problem in combinatorial optimization and has broad applications (see, e.g., \citep{krause2014submodular}).
Here we show how to formulate this classical combinatorial optimization problem as a submodular MDP.
We set the planning horizon $H = k$ and the state space $\states_h = \{s_h\}$.
The action set $\actions = \{1, 2, \ldots, n\}$.
The transition operator $\trans$ is deterministic and defined to be $\trans(s_h, a) = s_{h + 1}$ for any $a \in \actions$ and $h < k$.
Finally, we set the objective function
$f(\{(s_1, a_1), (s_2, a_2), \ldots, (s_H, a_H)\}) =\hat{f}(\{\omega_i \mid i \in \{a_1, a_2, \ldots, a_H\}\})$.
Clearly, since $\hat{f}$ is monotone and submodular, in this problem, $f$ is also monotone and submodular.

\paragraph{Log-determinant Objective Function.}
Here we consider an extension of the standard MDP setting, where the reward function $r(s, a) \in \mathbb{R}^{d \times d}$ returns a $d \times d$ positive semi-definite matrix instead of a real number.
The objective function $f$ is defined to be 
$
f(\{(s_1, a_1), (s_2, a_2), \ldots, (s_H, a_H)\})= \ln \det \left(\sum_{h = 1}^H r(s_h, a_h) + \lambda I\right)
$
 where $\lambda > 0 $ is a regularization term to avoid singular matrices. 
 When $d = 1$, the above setting degenerates to the standard MDP setting.
 As shown in~\citep{shamaiah2010greedy}, such an objective function is monotone and submodular, and thus can be covered by our framework.
 
We remark that it is possible to formulate a large family of applications as submodular MDPs using the log-determinant objective function.
For the measurement selection problem mentioned in the introduction, we may set $r(s, a)$ to be the Fisher information matrix associated with each state-action pair. In this case, maximizing the log-determinant objective function is equivalent to minimizing the volume of the error ellipsoid~\citep{kirchner2019heterogeneous}. 
In navigation tasks, if there are $d$ locations to be explored, we may set $r(s, a)$ to be a $d \times d$ diagonal matrix where the $i$-th diagonal entry denotes whether the $i$-th location is observable from state $s$.
In this case, maximizing the log-determinant objective function is equivalent to maximizing the {\em geometric mean} of the number of times each location is observed, which could have more favorable properties than maximizing the arithmetic mean.

\section{Continuous Greedy for Submodular MDPs}\label{sec:alg}
In this section, we present our algorithm for solving the planning problem in submodular MDPs.
We need the following notations.
\begin{itemize}
    \item For a vector $x \in [0, 1]^{\states \times \actions}$, let $S^x \subseteq \states \times \actions$ be the random subset of the state-action pairs, where each $e \in \states \times \actions$ appears in $S^x$ independently with probability $x_e$.
    \item For a random subset $R \subseteq \states \times \actions$, let $x^R$ denote the marginal probabilities that each state-action pair $e$ is in $R$, i.e., for each $e \in  \states \times \actions$,
    $
        x^R_e = \Pr_R[e \in R]
    $.
\end{itemize}

We first describe and analyze an idealized continuous version of the algorithm, and then present its discrete version, which is applicable in practice.

\subsection{Continuous Greedy}\label{sec:continuous}
\begin{definition}[Multilinear Extension]
    Given a nonnegative monotone submodular function $f: 2^{\states \times \actions} \to \mathbb{R}_+$, define its {\em multilinear extension} $F: [0, 1]^{\states \times \actions} \to \mathbb{R}_+$ as
    $
        F(x) = \sum_{S \subseteq \states \times \actions} f(S) \cdot \prod_{e \in S} x_e \cdot \prod_{e \notin S} (1 - x_e)
    $.
\end{definition}
In other words, $F(x) = \mathbb{E}_{S^x}[f(S^x)]$, where as defined above each $e \in \states \times \actions$ appears in $S^x$ independently with probability $x_e$.
Thus, one can approximately evaluate $F$ by sampling $S^x$ and taking the empirical mean.

In general, a random policy induces an exponentially large distribution over trajectories, which is impossible to represent efficiently.
We will use the multilinear extension $F$ as an approximation of the actual objective function, which guides the way we update the policy.
In fact, $F$ by a factor of $(1 - 1/e)$ upper bounds the actual objective function (cf. Lemma~\ref{lem:two-sided-boundedness}).
Moreover, $F$ by a factor of $H$ lower bounds the objective function where $H$ is the planning horizon (cf. Lemma~\ref{lem:two-sided-boundedness}).

Our plan is to (approximately) maximize the multilinear extension $F$ over the region corresponding to the marginal probabilities of all possible policies.
We now define this region.
Observe that each policy $\pi$ induces a distribution over trajectories.
Let $x(\pi) \in [0, 1]^{\states \times \actions}$ be the marginal probabilities that each state-action pair is in the random trajectory induced by $\pi$, i.e., for any $e \in \states \times \actions$,
$(x(\pi))_e = \Pr_{\pi\text{ induces trajectory }T}[e \in T]$.
Let $\mathcal{T}^* \subseteq [0, 1]^{\states \times \actions}$ be the set of vectors induced by all feasible policies, i.e.,
$
    \mathcal{T}^* = \{x(\pi) \mid \pi\text{ is a feasible policy}\}
$.
Moreover, let $\mathcal{T}$ be the downward closure of $\mathcal{T}^*$, i.e.,
$
    \mathcal{T} = \{x \ge 0 \mid \exists x' \in \mathcal{T}^*: x \le x'\}
$.
Note that $\mathcal{T}$ itself is downward closed, i.e., for $0 \le x \le x'$, if $x' \in \mathcal{T}$, then $x \in \mathcal{T}$.
We will maximize $F$ over $\mathcal{T}$.

Consider the following continuous greedy algorithm for maximizing $F$ on $\mathcal{T}$.
The algorithm defines a path $y: [0, 1] \to \mathcal{T}$, where $y(0) = 0 \in [0, 1]^{\states \times \actions}$ and $y(1)$ is the output of the algorithm.
In the continuous algorithm, $y$ is defined by a differential equation, and the derivative of $y$ (which defines $y$ on $(0, 1]$) is chosen greedily in $\mathcal{T}$ maximizing the growth of $F$, i.e.,
$\ddt y(t) = \operatorname{argmax}_{x \in \mathcal{T}} x \cdot \nabla F(y(t))$.

The following lemma can be proved using techniques in~\citep{calinescu2011maximizing}.
We include a proof in the appendix to help unfamiliar readers develop intuition.

\begin{lemma}\label{lem:continous_greedy}
    $y(1) \in \mathcal{T}$.
    Moreover, $F(y(1)) \ge (1 - 1 / e) \max_{x \in \mathcal{T}} F(x)$.
\end{lemma}
By the above lemma, $y(1)$ naturally defines a policy $\pi : \states \times \actions$, where for each $(s, a) \in \states \times \actions$, $\pi(a \mid s) =y(1)_{(s, a)} / \sum_{a' \in \actions}y(1)_{(s, a')}$.
In Section~\ref{sec:prac}, we discuss how to implement the above continuous algorithm practically.

\subsection{Practical Implementation}\label{sec:prac}

The above procedure provides some high-level guidance, but issues remain for solving submodular MDPs.
To implement the algorithm in a practical way, we need to answer the following questions.
\begin{itemize}
    \item The above procedure is continuous in its nature.
    How can we discretize the procedure, roughly preserving the approximation guarantee?
    \item How can we efficiently compute the gradient $\nabla F(y(t))$?
    \item For a given $\nabla F(y(t))$, how can we efficiently solve $\operatorname{argmax}_{x \in \mathcal{T}} x \cdot \nabla F(y(t))$?
    \item How can we recover a policy from the vector $y$?
\end{itemize}

\begin{figure*}[!t]
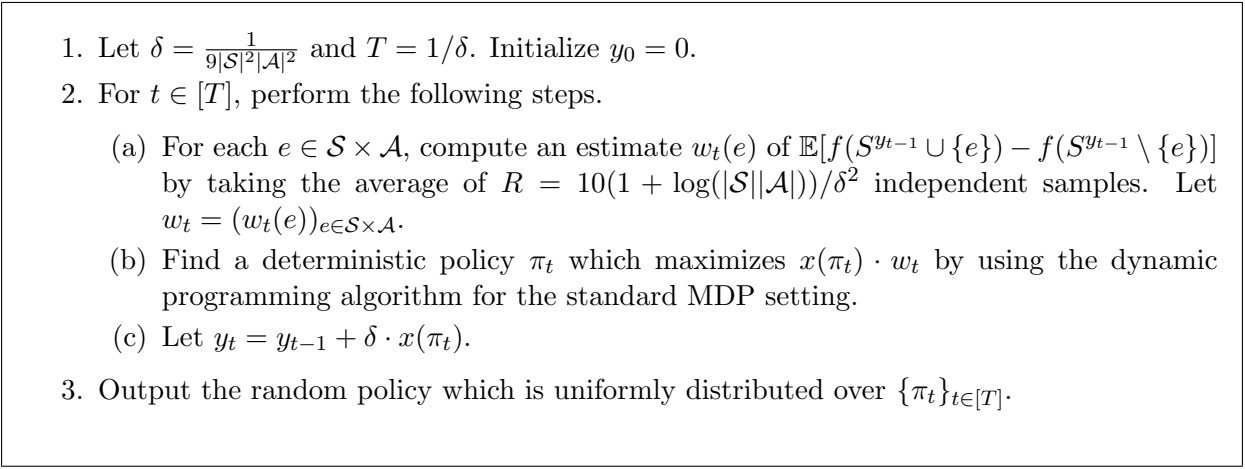

\begin{framed}
\begin{enumerate}
\setlength{\itemsep}{0pt}
\setlength{\parskip}{2pt}
    \item Let $\delta = \frac{1}{9|\states|^2|\actions|^2}$ and $T = 1 / \delta$.
    Initialize $y_0 = 0$.
    \item For $t \in [T]$, perform the following steps.
    \begin{enumerate}
    \setlength{\itemsep}{0pt}
        \item For each $e \in \states \times \actions$, compute an estimate $w_t(e)$ of $\mathbb{E}[f(S^{y_{t - 1}} \cup \{e\}) - f(S^{y_{t - 1}} \setminus \{e\})]$
         by taking the average of $R = 10(1 + \log(|\states||\actions|)) / \delta^2$ independent samples.
        Let $w_t = (w_t(e))_{e \in \states \times \actions}$.
        \item Find a deterministic policy $\pi_t$ which maximizes $x(\pi_t) \cdot w_t$ by using the dynamic programming algorithm for the standard MDP setting. \label{step:dp}
        \item Let $y_t = y_{t - 1} + \delta \cdot x(\pi_t)$.
    \end{enumerate}
    \item Output the random policy which is uniformly distributed over $\{\pi_t\}_{t \in [T]}$.
\end{enumerate}
\end{framed}
\caption{Continuous Greedy for submodular MDP.}
\label{fig:alg}
\end{figure*}

To address these issues, we present a practical version of the continuous greedy algorithm in Figure~\ref{fig:alg}.

\paragraph{Finite step size.}
Compared to the continuous version of the algorithm defined in Section~\ref{sec:continuous}, we use a finite step size of $\delta = \frac{1}{9|\states|^2|\actions|^2}$.
As will be shown in Lemma~\ref{lem:fractional-approximation}, such finite step size suffices for funding a good approximate solution.

\paragraph{Gradient oracle.} In order to evaluate the gradient $\nabla F$, we note that $F$ is a linear function with respect to any fixed input variable.
Therefore, in order to calculate the partial derivative of $F$ with respect to a specific input variable, we set that variable to be $0$ and $1$, and use the difference of values of $F$ as the gradient.
We also use Monte-Carlo to estimate the value of $F$.

\paragraph{Find optimal $x \in \mathcal{T}$.}
To solve $\operatorname{argmax}_{x \in \mathcal{T}} x \cdot \nabla F(y(t))$, note this is equivalent to finding a policy that maximizes the cumulative reward in the standard MDP setting, where for each state-action pair $e \in \states \times \actions$, the reward value $r(e)$ is defined by the gradient $\frac{\partial F}{\partial x_e}$.
Thus, one can readily solve $\operatorname{argmax}_{x \in \mathcal{T}} x \cdot \nabla F(y(t))$ by using the dynamic programming algorithm.

\paragraph{Recover policy.} Finally, we notice that the vector $y$ naturally defines a policy, which is a random policy uniformly distributed over all policies obtained by the dynamic programming algorithm.
In Section~\ref{sec:rounding}, we discuss possible approaches for rounding random polices to deterministic policies.

Now we give the theoretical guarantee of the algorithm in Figure~\ref{fig:alg}.
Using techniques in~\citep{calinescu2011maximizing}, we can show that policy produced by our algorithm approximately maximizes the multilinear extension $F$ even if finite step size and Monte-Carlo is adopted. 
Notice that the approximation ratio guarantee is similar to Lemma~\ref{lem:continous_greedy}.
\begin{lemma}
\label{lem:fractional-approximation}
    Let $y_T = \frac1T \sum_{t \in [T]} x(\pi_t) \in \mathcal{T}$.
    With probability at least $0.99$,
   $F(y_T) \ge (1 - 1 / e - o(1)) \cdot \max_{x \in \mathcal{T}} F(x)$.
\end{lemma}

We now show multilinear extension is a good approximation to the expected objective function.
The proof can be found in the appendix. 
\begin{lemma}
\label{lem:two-sided-boundedness}
    For any feasible policy $\pi$,
    $
        \frac{1}{H} F(x(\pi)) \le f(\pi) \le \frac{e}{e - 1}F(x(\pi))
    $
    where $H$ is the planning horizon of the MDP.
\end{lemma}

Lemma~\ref{lem:fractional-approximation}~and Lemma~\ref{lem:two-sided-boundedness} together imply the following guarantee of the continuous greedy algorithm. The proof can be found in the appendix. 

\begin{theorem}
    With probability at least $0.99$, we have
    $
        f(\pi) \ge \frac{(1 - 1/e - o(1))^2}{H} \cdot f(\pi^*)
    $.
\end{theorem}
We note that the failure probability of the algorithm can be reduced to an arbitrarily small constant
by independent repetitions and taking the best solution found among all repetitions.
\subsection{Rounding Policies} \label{sec:rounding}
In this section, we focus on the special case that the transition operator $\trans$ is deterministic. 
For deterministic polices, it can be seen that the multilinear extension $F$ equals the objective value of the policy
and thus the approximation guarantee in Lemma~\ref{lem:two-sided-boundedness} can be improved.
Therefore, intuitively, our algorithmic framework is more in favor of deterministic policies than random ones for deterministic environments.
When the transition operator is deterministic, it is possible to round the random policy obtained by the algorithm in Figure~\ref{fig:alg} to a deterministic policy.
As we will show later, such rounding approaches could lead to superior theoretical guarantee and practical performance.
Below we list two possible rounding approaches.

\paragraph{Policy with highest objective value (\textsf{HIGH}).}
Since the output of the algorithm is a random policy uniformly over $\{\pi_t\}_{t \in [T]}$ where each $\pi_t$ is a deterministic policy, we may round the random policy to a deterministic policy by picking the deterministic policy with highest objective value.


\paragraph{Rounding by sub-trajectories (\textsf{SUB}).}
Here we give another rounding method inspired by the pipage rounding technique by~\citep{ageev2004pipage}.
The method works directly on the marginal probabilities of state-action pairs.
As long as some state-action pair has a fractional marginal probability, we find two disjoint sub-trajectories (a sub-trajectory is a consecutive part of a trajectory) on which each state-action pair has strictly positive marginal probability.
We then try to move probability mass from all state-action pairs from one sub-trajectory to another, till some state-action pair has $0$ marginal probability.
Since there are two ways of performing this operation, in general we choose the one that results in a larger value of the multilinear extension.
Observe that each time we perfrom this operation, the number of state-action pairs with non-zero marginal probability decreases at least by $1$.
It follows that the rounding terminates in $|\states||\actions|$ steps, yielding a deterministic policy as desired.
We also present an efficient implementation of this rounding approach in the appendix. 

\subsection{Special Cases} \label{sec:special}

Now we discuss the two special cases in Section~\ref{sec:example}.
We show in both cases, our algorithm recovers classical algorithms designed for the respective cases and have stronger guarantees than in the general case.

\paragraph{Additive Reward Functions.}
When the objective function $f$ is additive, i.e., for any $S \subseteq \states \times \actions$, $f(S) = \sum_{e \in S} f(\{e\})$.
In such cases, submodular MDPs degenerates to standard MDPs with $r(s, a) = f(\{(s, a)\})$, which can be solved optimally by the classical dynamic programming algorithm.
As we show below, when the objective function is additive, the continuous greedy algorithm produces exactly the same solution as dynamic programming, which is provably optimal.

\begin{theorem}\label{thm:additive}
    When the objective function $f$ is additive, the continuous greedy algorithm outputs a deterministic policy with the maximum expected reward.
\end{theorem}

\paragraph{Cardinality Constraints.}
When applied to the submodular maximization with cardinality constraint problem introduced in Section~\ref{sec:example}, if rounding by sub-trajectories (cf.~Section~\ref{sec:rounding}) is employed, our algorithm in Figure~\ref{fig:alg} recovers the classical continuous greedy algorithm with pipage rounding~\citep{calinescu2011maximizing} and thus enjoys an approximation ratio of $1 - 1/e - o(1)$.
Below we demonstrate the equivalence and prove the approximation ratio guarantee. 

\begin{theorem}\label{thm:cardinality}
    Let $\pi$ be the output of Algorithm~\ref{fig:alg} on a submodular MDP corresponding to the submodular maximization with cardinality constraint problem.
    Rounding $\pi$ by sub-trajectories yields a deterministic policy $\pi'$, such that with probability $0.99$, the objective value of $\pi'$ is maximum up to a factor of $1 - 1/e - o(1)$.
\end{theorem}

\section{Experiments}\label{sec:exp}

\paragraph{Experiment Setup.}
All of our experiments are performed on the grid world environment. 
In our setting, the state space $\states = \{(i, j) \mid 1 \le i, j \le n\}$, where state $(i, j)$ corresponds to the grid in the $i$-th row and $j$-th column.
The action set $\actions = \{R, D\}$, where $R$ corresponds to moving right and $D$ corresponds to moving downward.
There are $H = 2n - 1$ levels of states, the initial state $s_1 = (1, 1)$ is the upper-left corner, and the final state $s_H = (n, n)$ is the lower-right corner. 
For each state-action pair $(s, a)$, there is a positive semi-definite (PSD) matrix $r(s, a) \in \mathbb{R}^{d \times d}$ associated with that state-action pair. 
In our experiments, for simplicity, $r(s, a)$ are all diagonal matrices with non-negative diagonal entries and the size of diagonal matrices $d = 10$.
The goal is to find a policy to maximize the objective function
$f(r_1, r_2, \ldots, r_H) = \ln \det (\sum_{h = 1}^H r_h + \lambda I)$, where $r_h = r(s_h, a_h)$ and $\lambda = 10^{-5}$ is a regularization term.
Notice that the objective function defined above is the log-determinant objective function mentioned in Section~\ref{sec:example}, and therefore, the problem defined above is a submodular MDP.
In our experiments, we work with two different sets of reward functions $r$: synthetic environment and a navigation task.

\paragraph{Synthetic Environment.} 
In the synthetic environment, the size of the grid $n = 10$ or $n = 20$. 
For each $(s, a) \in \states \times \actions$, $r(s, a)$ is a diagonal matrix whose first $d / 2$ diagonal entries are drawn uniformly and independently from $\{0, 1, \ldots, 10\}$, and the rest $d/2$ diagonal entries are $0$. 
Then, for each $i \in \{d / 2 + 1, d/ 2 + 2, \ldots, d\}$, we randomly choose $t$ different states $s^{(i, 1)}, s^{(i, 2)}, \ldots, s^{(i, t)} \in \states$ and $t$ actions $a^{(i, 1)}, a^{(i, 2)}, \ldots, a^{(i, t)} \in \actions$, and for each $1 \le j \le t$, $r(s^{(i, j)}, a^{(i, j)})$ is replaced with a matrix whose $(i, i)$-th entry is $1$ and all other entries are $0$.
In our experiments, we set $t = 2$ or $t = 5$.
Since $t$ is small, for this task, intuitively there are two different kinds of diagonal entries: ``uniform'' diagonal entries that correspond to the first $d / 2$ diagonal entries, and ``sparse'' diagonal entries that correspond to the rest diagonal entries. 
To achieve a higher objective value,
the agent must visit at least one state-action pair with non-zero value for each ``sparse'' diagonal entry, while carefully choosing the policy to maximize the objective value for the first $d / 2$ ``uniform'' diagonal entries. 

\begin{figure}[t]
    \centering
    \begin{subfigure}
        \centering
        \includegraphics[width=0.23\textwidth,interpolate=false]{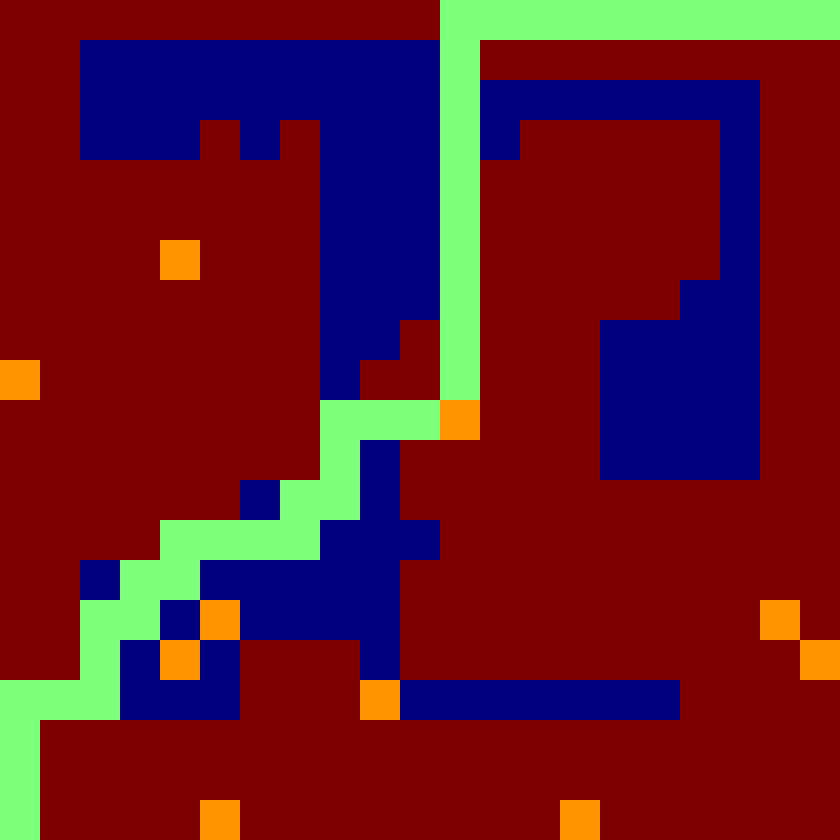}
    \end{subfigure}%
    ~
    \begin{subfigure}
        \centering
        \includegraphics[width=0.23\textwidth,interpolate=false]{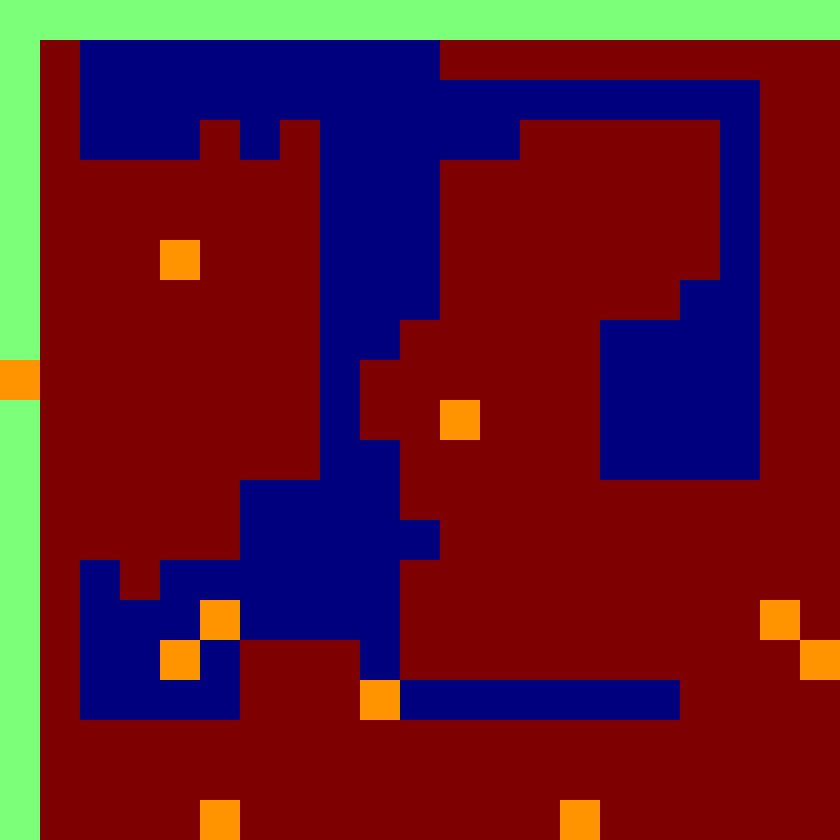}
    \end{subfigure}
    ~
    \begin{subfigure}
        \centering
        \includegraphics[width=0.23\textwidth,interpolate=false]{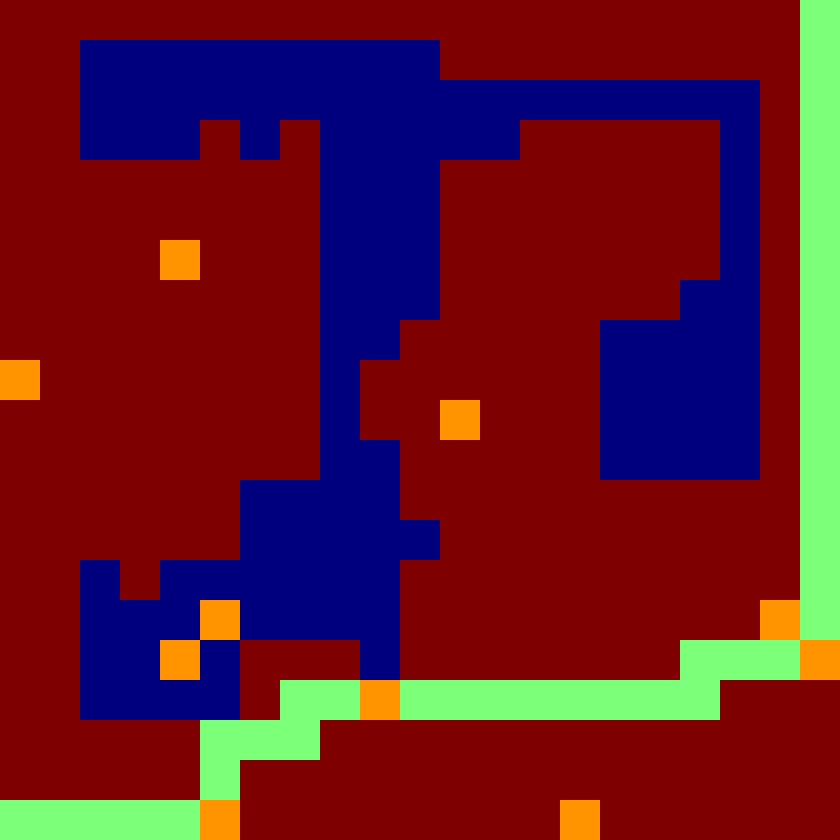}
    \end{subfigure}
    \caption{ Navigation task $\mathsf{Nav}_1$.
    Obstacles are marked red, navigable locations are marked blue, and locations to be explored are marked orange. 
    Trajectories produced by different algorithms are marked green.
    Left: trajectory produced by DP in $\mathsf{Aug}_1$. 
    Middle: trajectory produced by greedy in $\mathsf{Aug}_1$. 
    Right: trajectory produced by continuous greedy with $\delta = 0.01$ and $R = 10$, together with \textsf{HIGH} rounding.
    The goal is to the maximize geometric mean of the number of times each orange location is visited, and the trajectory on the right (produced by our algorithm) achieves a better objective value since it visits more orange locations than the other two trajectories. 
    }    
    \label{fig:nav}
\end{figure}
\paragraph{Navigation Task.}
The navigation task is based on real-world scans from the Matterport Dataset by~\citep{chang2017matterport3d}. 
We create three maps (named as $\mathsf{Nav}_1$, $\mathsf{Nav}_2$ and $\mathsf{Nav}_3$) of size $21 \times 21$ ($n=21$) from Matterport reconstructions by computing the obstacles and navigable locations.
Each square cell in the map has a side length of $20$cm in the real-world. 
For each map, we randomly choose $d=10$ locations from the map and designate them as locations to be explored.
Consequently, $r(s,a)$ is a $d \times d$ diagonal matrix where the $i$-th diagonal entry denotes if location $i$ is visible from state $s$. 
Note that by maximizing the objective function defined above, we are effectively maximizing the {\em geometric mean} of the number of times each location is visited.
In our experiments, visibility is defined as follows: a cell $j$ is visible from another cell $i$ if all the cells connecting $i$ and $j$ are navigable and the distance between the two cells is less than the vision range, or cell $j$ is adjacent to cell $i$.
In our experiments we set the vision range to be $3$. 
See Figure~\ref{fig:nav} for a visualization of the the navigation task $\mathsf{Nav}_1$.

\paragraph{Algorithms and Baselines.}
In our experiments, we implement the algorithm in Figure~\ref{fig:alg}.
Instead of using the theoretical bound for the step size $\delta$ and the number independent samples $R$, we treat $\delta$ and $R$ as tunable parameters and study their effects on the objective value.
Since the final policy of the algorithm is a random policy uniformly chosen from $T$ deterministic policies, we report the mean objective value of the $T$ deterministic policies which is the expected objective value of the output policy. 
We implement the two rounding approaches (\textsf{HIGH} and \textsf{SUB}) in Section~\ref{sec:rounding} and report the objective value of the resulting policies.

We compare our algorithm with two baselines: dynamic programming (DP) and greedy.
For DP, submodular MDP is treated as a standard MDP where the reward value for a state-action pair is defined to be $\ln\det(r(s, a) + \lambda I)$.
Note that such reward value is equal to 
$
f(\{(s_1, a_1), \ldots, (s_h, a_h)\}) 
- f(\{(s_1, a_1),  \ldots, (s_{h - 1}, a_{h - 1})\})
$
when $r(s_1, a_1) = \ldots = r(s_{h - 1}, a_{h - 1}) = 0$ (up to a constant term).\footnote{We note that in general, it is impossible to design a reward function which always equals $f(\{(s_1, a_1), \ldots, (s_h, a_h)\}) - f(\{(s_1, a_1), \ldots, (s_{h - 1}, a_{h - 1})\})$. 
}
Therefore, DP in such an MDP is a natural baseline algorithm to compare with. 
For the greedy algorithm, for each step, the agent greedily chooses an action so that the objective value is maximized after that step.
We further apply DP and greedy in MDP with augmented action space. 
In the MDP $\mathsf{Aug}_l$ which has $l$ levels of augmentation, in each step, the agent chooses $l$ actions for $l$ consecutive steps in the original MDP. 
Therefore, in each step in $\mathsf{Aug}_l$, the agent receives a reward value that corresponds to $l$ consecutive steps in the original MDP, and apply the transition operator in the original MDP for $l$ times. 
With larger values for $l$, the agent takes more global information into consideration and is expected to output a better policy. 
Since the action space has size $|\actions|^l$ in $\mathsf{Aug}_l$, we set $l$ to be $1$, $2$ or $3$ for efficiency considerations. 

We also compare our algorithm with a strong baseline based on Deep $Q$-Networks (DQN)~\citep{mnih2015human}.
To motivate such a baseline, we note that the any submodular MDP can be casted as a standard MDP with augmented state space.
In the state-augmented MDP, each state in level $h$ is of the form $(s_1, a_1,\ldots, s_h, a_h)$, where $(s_1, a_1), \ldots, (s_h, a_h)$ are all the state-action pairs in the first $h$ steps on a trajectory in the original MDP. 
In the state-augmented MDP, we define a new reward function
$
r'(s_1, a_1, \ldots, s_h, a_h) = f(\{(s_1, a_1), \ldots, (s_h, a_h)\}) - f(\{(s_1, a_1), \ldots, (s_{h - 1}, a_{h - 1})\}).
$
The size of the state space in the state-augmented is $(|\states| |\actions|)^H$, which is unacceptable in practice. 
We note that in environments considered above, we may compactly represent each state $(s_1, a_1, \ldots, s_h, a_h)$ in the state-augmented as 
$
\sum_{i = 1}^h r(s_i, a_i) \in \mathbb{R}^{d \times d}
$ since $r'$ depends only on the sum of the reward matrices. 
Given such a representation, we may now use DQN to solve problems considered in our experiments as a baseline. 
Hyperparameters used in DQN are provided in the appendix.


\setlength{\tabcolsep}{2pt}
\begin{table*}[!t]
\centering
\footnotesize
\begin{tabular}{l|llll|lll}
\hline
Algorithm & $\mathsf{Syn}_{10, 2}$ & $\mathsf{Syn}_{10, 5}$ & $\mathsf{Syn}_{20, 2}$ & $\mathsf{Syn}_{20, 5}$ & $\mathsf{Nav}_1$ & $\mathsf{Nav}_2$ &$\mathsf{Nav}_3$ \\ 
\hline
DP in $\mathsf{Aug}_1$        			& $-34.7$ $(0.3)$    & $-34.8$ $(0.2)$    & $-31.0$ $(0.2)$    & $-31.0$ $(0.1)$   & $-41.4$ $(14.9)$ & $-42.0$ $(15.1)$ & $-47.2$ $(13.8)$ \\
DP in $\mathsf{Aug}_2$        			& $-3.4$ $(8.7)$      & $7.1$ $(7.8)$       & $4.4$ $(8.8)$       & $13.7$ $(8.9)$     & $-38.3$ $(14.3)$ & $-40.1$ $(14.3)$ & $-46.1$ $(13.7)$ \\
DP in $\mathsf{Aug}_3$        			& $3.3$ $(8.2)$       & $13.4$ $(7.3)$     & $9.8$ $(7.5)$        & $18.3$ $(7.9)$    & $-36.4$ $(14.2)$ & $-38.6$ $(14.7)$ & $-45.1$ $(14.0)$ \\
Greedy in $\mathsf{Aug}_1$ 			& $-14.3$ $(11.9)$  & $1.3$ $(11.2)$      & $-20.7$ $(9.9)$    & $-7.5$ $(13.1)$   & $-85.9$ $(16.4)$ & $-95.5$ $(13.9)$ & $-92.9$ $(14.9)$ \\
Greedy in $\mathsf{Aug}_2$ 			& $-8.8$ $(12.2)$    & $6.9$ $(10.1)$      & $-16.7$ $(11.8)$  & $-1.0$ $(13.9)$   & $-80.0$ $(17.9)$ & $-84.4$ $(18.7)$ & $-87.7$ $(18.1)$ \\
Greedy in $\mathsf{Aug}_3$ 			& $-5.0$ $(11.6)$    & $11.9$ $(8.7)$       & $-12.2$ $(10.7)$  & $2.5$ $(12.3)$    & $-72.3$ $(21.1)$ & $-73.3$ $(20.4)$ & $-83.5$ $(19.3)$ \\
DQN                                      			& $1.8$ $(7.6)$       & $16.7$ $(5.8)$      & $-0.0$ $(7.0)$      & $13.3$ $(6.7)$     & $-35.4$ $(13.3)$ & $-35.3$ $(11.9)$ & $-46.4$ $(14.0)$\\
\hline\hline
$\mathsf{CG}_{0.1, 10}$     			& $-11.6$ $(5.1)$    & $-0.9$ $(4.1)$        & $-11.5$ $(4.5)$    & $-1.9$ $(4.6)$      & $-58.7$ $(9.4)$ &$-58.0$ $(10.2)$ & $-64.9$ $(10.4)$ \\
$\mathsf{CG}_{0.1, 10}$ + \textsf{HIGH}   & $4.9$ $(7.3)$       & $16.2$ $(5.9)$       & $7.3$ $(8.3)$       & $17.2$ $(7.0)$      & $-36.2$ $(12.8)$ & $-38.3$ $(13.4)$ & $-45.1$ $(13.4)$ \\
$\mathsf{CG}_{0.1, 10}$ + \textsf{SUB}    & $4.6$ $(7.5)$        & $16.4$ $(6.6)$       & $7.9$ $(8.3)$       & $18.5$ $(7.6)$      & $-39.3$ $(14.5)$ & $-41.3$ $(14.5)$ & $-49.1$ $(14.6)$ \\
\hline\hline
$\mathsf{CG}_{0.01, 10}$ 				& $-11.8$ $(3.8)$    & $-0.3$ $(3.1)$        & $-12.1$ $(3.7)$     & $-2.4$ $(3.3)$      & $-58.9$ $(9.5)$& $-59.2$ $(9.6)$ & $-66.1$ $(10.2)$\\
$\mathsf{CG}_{0.01, 10}$ + \textsf{HIGH} & $\bf{8.2}$ $(7.1)$ & $\bf{20.7}$ $(3.0)$ & $11.6$ $(7.4)$      & $23.6$ $(4.7)$      & $\bf{-32.8}$ $(12.9)$ & $\bf{-32.9}$ $(12.1)$ & $\bf{-41.9}$ $(13.9)$ \\
$\mathsf{CG}_{0.01, 10}$ + \textsf{SUB}  & $4.1$ $(7.9)$        & $18.1$ $(5.2)$       &$7.7$ $(8.2)$          & $20.2$ $(6.6)$     & $-37.8$ $(13.8)$ & $-38.0$ $(14.7)$ & $-47.0$ $(14.5)$ \\
\hline\hline
$\mathsf{CG}_{0.1, 100}$ 				& $-1.9$ $(5.7)$     & $7.2$ $(4.7)$         & $3.6$ $(6.7)$          & $11.9$ $(6.1)$      &$-50.8$ $(10.0)$ & $-51.8$ $(10.5)$ & $-56.2$ $(11.6)$\\
$\mathsf{CG}_{0.1, 100}$ + \textsf{HIGH} & $8.0$ $(6.9)$      & $19.8$ $(4.2)$       & ${\bf 12.5}$ $(7.3)$ & $23.5$ $(5.3)$       & $-34.0$ $(12.9)$ & $-35.4$ $(13.2)$ & $-43.5$ $(13.9)$ \\
$\mathsf{CG}_{0.1, 100}$ + \textsf{SUB}  &$7.7$ $(6.9)$       & $18.9$ $(5.1)$       & $12.1$ $(7.4)$         & $\bf{23.7}$ $(5.4)$ & $-36.2$ $(14.2)$ & $-37.1$ $(14.6)$ & $-45.3$ $(14.6)$ \\
\hline
\end{tabular}
\caption{Experiment Results. 
$\mathsf{Aug}_i$: Augmented MDP with $i$ levels of augmentation. 
$\mathsf{CG}_{\delta, R}$ : Continuous greedy with step size $\delta$ and $R$ independent samples. 
$\mathsf{Syn}_{n, t}$: Synthetic environment with grid size $n$ and parameter $t$.
$\mathsf{Nav}_i$: $i$-th navigation task. 
\textsf{HIGH} and \textsf{SUB} are the rounding approaches defined in Section~\ref{sec:rounding}. For each entry in the table, the first number is the mean objective value and the second number is the standard deviation of the objective value, among $100$ repetitions. }
\label{tab:result}
\end{table*}

\paragraph{Results and Discussions.}
We report the experiment results in Table~\ref{tab:result}.
For each environment we repeat the experiments for $100$ times and report the mean and the standard deviation of the objective value.
We also provide more detailed comparison between the baseline algorithms and our algorithm in the appendix.
Here we make a few observations regarding the results.
First, our approach significantly outperforms baseline algorithms.
This is because DP and greedy find polices by locally maximizing the objective function while our approach takes more global information into consideration. 
Moreover, DQN cannot exploit the submodularity of the underlying problem as our algorithm.  
Secondly, employing rounding approaches improves the performance of our algorithm.
As mentioned in Section~\ref{sec:rounding}, for deterministic polices, the multilinear extension equals the objective value of the policy, and thus the approximation guarantee is improved.
Our experiment results verify the intuition that our algorithmic framework is more in favor of deterministic policies than random ones.
Finally, smaller step size and larger sampling size improve the performance. 
This is because smaller step size gives a better approximation of the continuous greedy procedure defined in Section~\ref{sec:continuous}, and larger sampling size gives a better estimate of the multilinear extension.

\section{Conclusion}\label{sec:conclusion}
In this paper, we study planning with submodular objective functions.
We propose a theoretically principled algorithmic framework which recovers classical algorithms when applied to standard planning and submodular maximization with cardinality constraints. 
In our experiments, our approach outperforms baseline algorithms, which demonstrates the practicality of our approach.
An interesting future direction is to generalize our algorithmic framework to the setting where the state space is large and thus one needs to combine function approximation techniques with our approach. 

\section*{Acknowledgements}
R. Wang, D. S. Chaplot and R. Salakhutdinov are supported in part by DARPA HT0011990016 and NSF IIS1763562. 
H. Zhang is supported in part by  NSF IIS1814056. 
The views, opinions and/or findings expressed are those of the author and should not be interpreted as representing the official views or policies of the Department of Defense or the U.S. Government.
%
%

\bibliography{ref}
\bibliographystyle{abbrvnat}

\onecolumn
\appendix
\section{Missing Proofs}
We will need the following properties of the multilinear extension in our proofs.
\begin{lemma}[\citep{calinescu2011maximizing}]
    For any monotone submodular $f$, its multilinear extension $F$ satisfies:
    \begin{itemize}
        \item $F$ is monotone, i.e., for any $x \le y$, we have $F(x) \le F(y)$.
        \item For any $x \in [0, 1]^{\states \times \actions}$, $y \ge 0$, the univariate function $F(x + \xi \cdot y)$ is concave in $\xi$.
    \end{itemize}
\end{lemma}
\subsection{Proof of Lemma~\ref{lem:continous_greedy}}
\begin{proof}
    First observe that $y(1)$ is a convex combination of vectors in $\mathcal{T}$, so $y(1) \in \mathcal{T}$.
    Now we prove the second part of the claim.
    Let $x^* = \operatorname{argmax}_{x \in \mathcal{T}} F(x)$, and $\opt = F(x^*)$.
    The goal is to show by construction that for any $t \in [0, 1]$,
    \[
        \ddt F(y(t)) = \max_{x \in \mathcal{T}} x \cdot \nabla F(y(t)) \ge \opt - F(y(t)).
    \]
    Elementary calculus then gives \[F(y(1)) \ge (1 - 1 / e) \opt.\]
    We show that \[v(t) = \max(x^* - y(t), 0)\] satisfies
    \[
        v(t) \cdot \nabla F(y(t)) \ge \opt - F(y(t)).
    \]
    Observe that $v(t) \le x^* \in \mathcal{T}$ and $\mathcal{T}$ is downward closed, so $v(t) \in \mathcal{T}$ for any $t$.
    Consider \[g_t(\xi) = F(y(t) + \xi \cdot v(t))\] as a function of $\xi$.
    $g_t$ is concave in $\xi$, since $v(t) \ge 0$.
    Moreover, \[y(t) + v(t) = \max(x^*, y(t)) \ge x^*,\] so by the monotonicity and concavity of $F$,
    \begin{align*}
        & v(t) \cdot \nabla F(y(t)) \\
         = &\left.\frac{g_t(\xi)}{\mathrm{d}\xi}\right|_{\xi = 0} \ge g_t(1) - g_t(0) \tag{concavity of $g$} \\
         = &F(y(t) + v(t)) - F(y(t)) \ge F(x^*) - F(y(t)) \tag{monotonicity of $F$} \\
         = &\opt - F(y(t)).
    \end{align*}
    This concludes the proof.
\end{proof}
\subsection{Proof of Lemma~\ref{lem:two-sided-boundedness}}
\begin{proof}
    Let $T$ be a random trajectory induced by $\pi$.
    The upper bound follows from Theorem~1 in~\citep{agrawal2011optimization}, which essentially says that for a random set $R \subseteq \states \times \actions$ with marginal probabilities $x^R$ on state-action pairs, it is always true that
    $\mathbb{E}_R[f(R)] \le \frac{e}{e - 1}F(x^R)$.
    Using this result, we immediately get the upper bound on $f(\pi)$.

    For the lower bound, on one hand, we have
    \begin{align*}
        \mathbb{E}_P[f(P)] & = \sum_{P'\text{ is a trajectory}} \Pr[\pi\text{ induces }P'] \cdot f(P') \\
        & \ge \sum_{P'} \Pr[\pi\text{ induces }P'] \cdot \max_{e \in P'} f(\{e\}) \tag{monotonicity of $f$} \\
        & \ge \sum_{P'} \Pr[\pi\text{ induces }P'] \cdot \frac{1}{|P|} \sum_{e \in P'} f(\{e\}) \\
        & = \sum_{P'} \Pr[\pi\text{ induces }P'] \cdot \frac{1}{H} \sum_{e \in P'} f(\{e\}) \\
        & = \frac{1}{H} \sum_{e \in \states \times \actions} x(\pi)_e f(\{e\}).
    \end{align*}
    On the other hand,
    \begin{align*}
        \mathbb{E}_{S^{x(\pi)}}[f(S^{x(\pi)})] & = \sum_{S \subseteq \states \times \actions} \Pr[S^{x(\pi)} = S] \cdot f(S) \\
        & \le \sum_{S \subseteq \states \times \actions} \Pr[S^{x(\pi)} = S] \cdot \sum_{e \in S} f(\{e\}) \tag{submodularity of $f$} \\
        & = \sum_{e \in \states \times \actions} x(\pi)_e f(\{e\}).
    \end{align*}
    It follows that
    \[
        \frac{1}{H} F(x(\pi)) \le f(\pi),
    \]
    which concludes the proof.
\end{proof}

\subsection{Proof of Theorem~\ref{thm:additive}}
\begin{proof}
    Let $F$ be the multilinear extension of $f$.
    The key observation is that $F$ is linear whenever $f$ is additive.
    In fact, for $x \in [0, 1]^{\states \times \actions}$, $F(x) = \mathbb{E}_{S^x}[f(S^x)] =  \sum_{e \in \states \times \actions} x_e \cdot f(\{e\})$.

    As a result, $\nabla F(y_t)$ is independent of $y_t$, and is simply $(f(\{e\}))_e$.
    The dynamic programming algorithm in Step~\ref{step:dp} in Figure~\ref{fig:alg} therefore always finds the same policy $\pi$ in each iteration, which is precisely the optimal policy $\pi^*$ with respect to the addtive objective function $f$.
    The convex combination of all these same policies computed in each iteration is clearly the same policy.
    The optimality follows.
\end{proof}

\subsection{Proof of Theorem~\ref{thm:cardinality}}
\begin{proof}
When applying rounding by sub-trajectories on the submodular MDP corresponding to the submodular maximization with cardinality constraint problem, the algorithm will check all levels $h \in \{1, 2, \ldots, k\}$ one by one.
For each level $h$, if there are two actions $a_1, a_2 \in \{1, 2, \ldots, n\}$ such that both actions have strictly positive marginal probabilities, then the algorithm will move probability mass from one to another to maximize the value of the multilinear extension. Notice that this is equivalent to the standard pipage rounding algorithm~\citep{ageev2004pipage}.

In order to prove the approximation ratio guarantee, notice that since the objective function is submodular, among the two ways to move probability mass, there is at least one way so that the value of the multilinear extension does not decrease after moving.
Therefore, throughout the algorithm, the value of the multilinear extension never decreases.
By Lemma~\ref{lem:fractional-approximation}, the discretized continuous greedy algorithm finds a solutions with approximation ratio $1 - 1/e - o(1)$ with respect to the multilinear extension.
Moreover, after the rounding procedure the policy is deterministic, and for deterministic policies the value of the multilinear extension equals the objective function value.
Thus, the algorithm returns a solution with approximation ratio $1 - 1/e - o(1)$ to the submodular maximization with cardinality constraint problem.  
\end{proof}

\section{Efficient Implementation of Rounding by Sub-Trajectories}

To efficiently implement the rounding by sub-trajectories approach, one may follow the steps below.
\begin{enumerate}
    \item Start from the initial state, and move forward in the MDP as long as the policy gives a deterministic action at the current state.
    \item After the previous step, now we have reached a state where the policy chooses a random action. 
    We then pick any two of these actions, and move forward by playing each of them respectively.
    The two actions lead to different states, creating two trajectories both starting from the current state.
    From now on, we keep track of the two sub-trajectories, till they converge again or reach the last level.
    \item As long as the two sub-trajectories do not overlap (except at the starting point), pick any action played by the policy with positive probability for both sub-trajectories, and move forward in the MDP one step at a time for the two sub-trajectories simultaneously.
    \item At some point the two sub-trajectories converge or reach the last level.
    We terminate the search immediately, and perform the rounding operation for the two sub-trajectories found.
\end{enumerate}
\section{More Experimental Results}\label{sec:more_exp}
\begin{figure}
\centering
\includegraphics[scale=0.71]{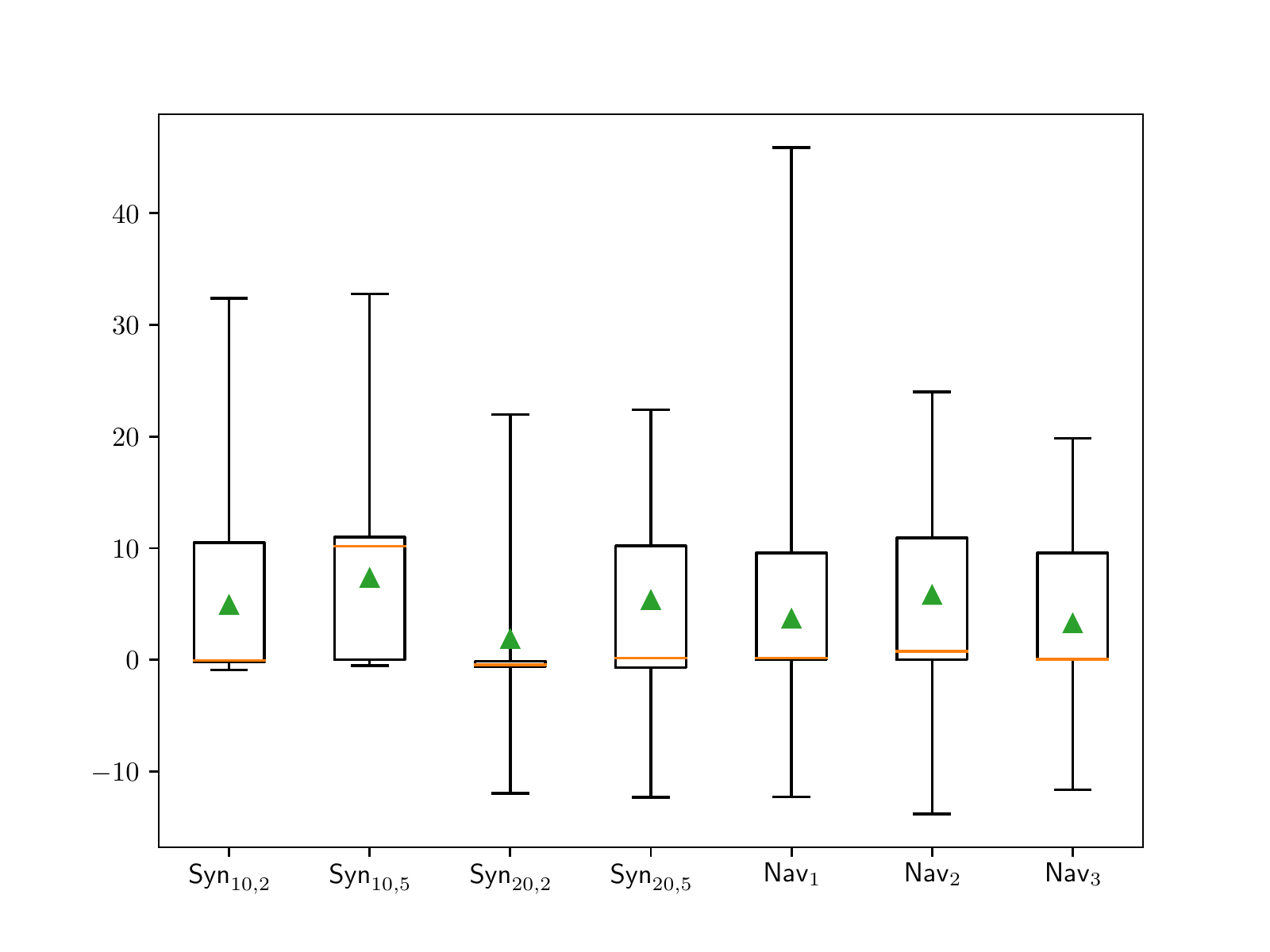}
\caption{Box plots of performance difference between $\mathsf{CG}_{0.01, 10}$ + \textsf{HIGH} and DP in $\mathsf{Aug}_3$ in different environments. For each of the $100$ repetitions, for each of the $7$ environments, we calculate the difference of the objective function of the policy returned by $\mathsf{CG}_{0.01, 10}$ + \textsf{HIGH} and that returned by DP in $\mathsf{Aug}_3$. Here the green triangle denotes the mean and the orange line denotes the median. }
\label{fig:more_exp_1}
\end{figure}

\begin{figure}
\centering
\includegraphics[scale=0.71]{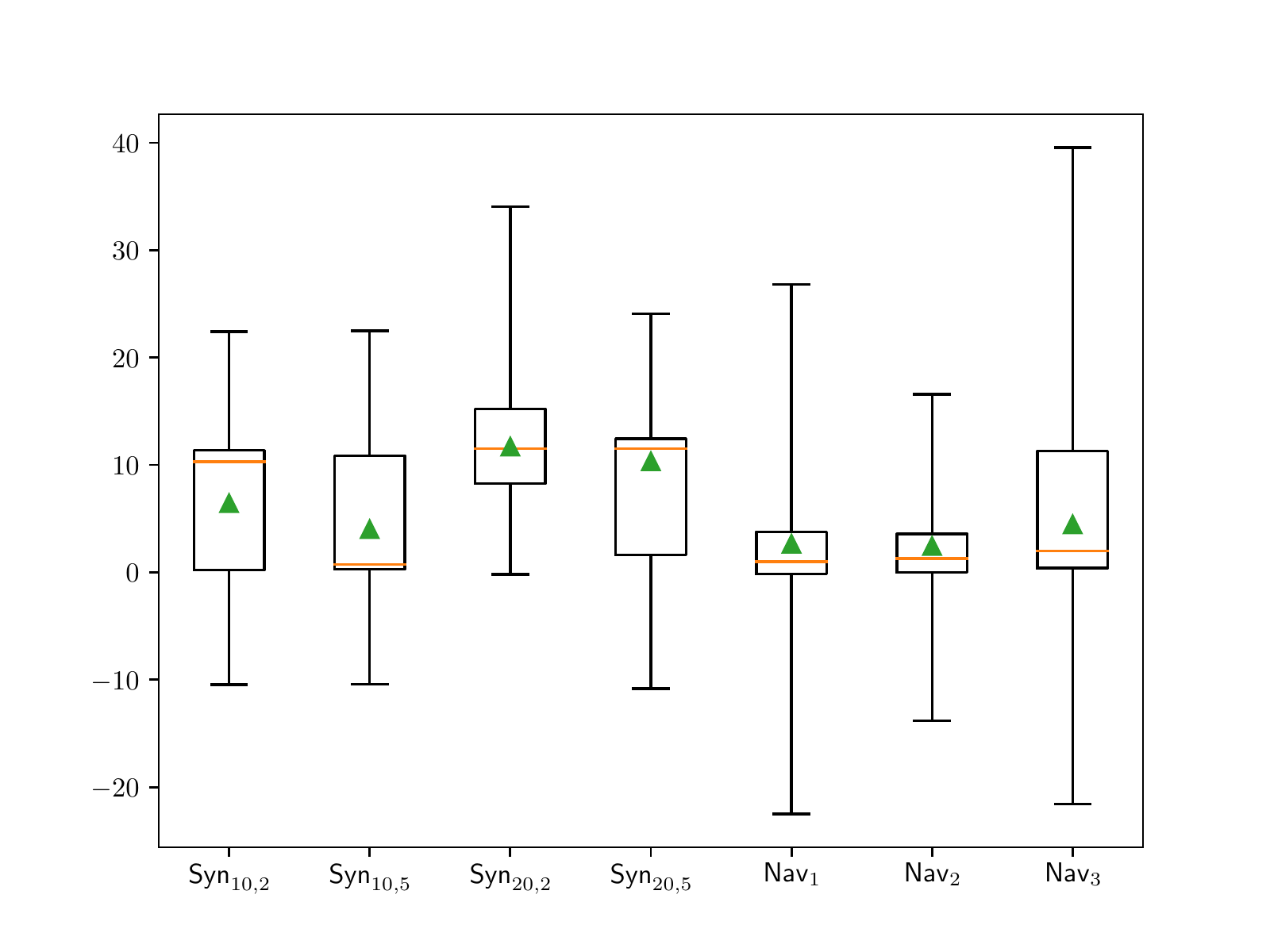}
\caption{Box plots of performance difference between $\mathsf{CG}_{0.01, 10}$ + \textsf{HIGH} and DQN in different environments. For each of the $100$ repetitions, for each of the $7$ environments, we calculate the difference of the objective function of the policy returned by $\mathsf{CG}_{0.01, 10}$ + \textsf{HIGH} and that returned by DQN. Here the green triangle denotes the mean and the orange line denotes the median. }
\label{fig:more_exp_2}
\end{figure}

In this section, we provide a more detailed comparison between two baseline algorithms DP in $\mathsf{Aug}_3$ and DQN, and our algorithm $\mathsf{CG}_{0.01, 10}$ + \textsf{HIGH} in Figure~\ref{fig:more_exp_1} and Figure~\ref{fig:more_exp_2}.
From the results in Figure~\ref{fig:more_exp_1} and Figure~\ref{fig:more_exp_2}, it can be seen that $\mathsf{CG}_{0.01, 10}$ + \textsf{HIGH} performs no worth than both baseline algorithms for 75\% of the repetitions, which clearly demonstrates the effectiveness of our algorithm. 
\section{Hyperparameters in DQN}\label{sec:hyper}
Since in our experiments all reward matrices are $d \times d$ diagonal matrices, we take the diagonal entries of the sum of reward matrices $\sum_{i = 1}^h r(s_i, a_i) $ (which is a $d$-dimensional vector) as the input to the DQN.
There are three hidden layers in the DQN, each with $64$ hidden units.
We adopt the ReLU function as the activation function. 
The size of the replay buffer is set to be $10, 000$.
The batch size is set to be $64$.
The discounting factor is set to be $0.99$.
We update the target network every $20$ episodes.
The $\varepsilon$ parameter in $\varepsilon$-greedy for the $h$-th episode is set to be $0.05 + 0.95e^{-h/20}$.
The total number of episodes is set to be $200$.\footnote{The number of episodes is chosen so that the running time of DQN and our algorithms is roughly the same. }
We report the trajectory with largest objective value found by DQN during its execution. 
\end{document}